# Prompt-oriented Output of Culture-Specific Items in Translated African Poetry by Large Language Model: An Initial Multi-layered Tabular Review.

Adeyola Opaluwah
PhD Student
University of Antwerp, Belgium
adeyola.opaluwah@student.uantwerpen.be

**Abstract**

This paper examines the output of culture-specific items (CSIs) generated by ChatGPT pro in response to three structured prompts to translate three anthologies of African poetry. The first prompt was broad, the second focused on poetic structure, and the third emphasized cultural specificity. To support this analysis, four comparative tables were created. The first table presents the results of the CSIs produced after the three prompts; the second categorizes these outputs based on Aixelá's framework of "Proper nouns and Common expressions" (1996, p. 59); the third summarizes the CSIs generated by human translators, a custom-built translation engine (CTE), and a Large Language Model (LLM). The final table outlines the strategies employed by ChatGPT Pro following the culture-specific prompt. Compared to the outputs of CSIs from reference human translation (HT) and the CTE in prior studies, the findings indicate that the culture-oriented prompts used with ChatGPT Pro did not yield significant enhancements in the CSIs during the translation of African poetry from English to French. Among the 54 CSIs, the reference HT produced 33 CSIs in repetition, the CTE generated 38 CSIs in repetition, while ChatGPT pro produced 41 CSIs in repetition. The untranslated CSIs revealed inconsistencies in LLM's approach to translating CSIs in African poetry from English to French; some of the translations of the remaining 13 CSIs were inaccurate when compared to HT and CTE. This further supports previous research conclusions that the training data for LLMs have not adequately covered diverse languages and cultures, thus indicating that they are not yet ready to replace human translators.

## Introduction

Large Language Models (LLMs) are AI systems trained on vast amounts of text data to understand and generate human-like language. Utilizing neural networks and machine learning algorithms, LLMs have transformed natural language processing (NLP) by enabling accurate and efficient language comprehension and generation. These models are trained on diverse text sources such as Wikipedia, news, books, and social media, allowing them to identify linguistic patterns and relationships. This capability makes LLMs versatile, supporting applications like translation, chatbots, text summarization, and sentiment analysis. A significant advantage of LLMs is their ability to perform tasks without task-specific training, making them adaptable across fields like finance, healthcare, and education (Ray, 2023)

This paper's investigation extends prior research that analyzed the translation of Culture-specific items (CSIs) from English to French by General Machine Translation (GMT) and a custom machine translation engine (CTE) in three African anthologies. It evaluates explicitly how ChatGPT, a Large Language Model (LLM), translates CSIs in the same anthologies of African poetry authored by Wole Soyinka. The study employs three different prompts: the first prompt is a general translation request into French; the second prompt focuses on maintaining the poem's structure by preserving the stanzas; and the third is culture-oriented, asking for a translation that emphasizes cultural vocabulary. Initially, ChatGPT could not translate an entire anthology from English to French, leading to an upgrade to ChatGPT Pro after receiving notifications that only a few poems could be translated daily. It was believed that the Pro version would facilitate the translation of more poems, which prompted the request for a complete anthology translation. The response received from ChatGPT Pro is presented below.

"*To translate an anthology into French, you would typically need to follow these steps:*

1. ***Select the Text****: Choose the specific excerpts or poems you want to translate.*
2. ***Context Understanding****: Understand the themes and nuances of the original text, as literary translation often requires capturing tone and style.*
3. ***Translation****: Translate the text into French, paying attention to vocabulary, grammar, and idiomatic expressions.*
4. ***Review and Edit****: Revisit your translation for clarity and accuracy. It might be helpful to have a native speaker or professional translator review it.*

5. ***Finalize**: Format the translated anthology appropriately, ensuring it's accessible to readers.*

*If you have a specific excerpt or text from the anthology you would like translated, feel free to share it!*"

Previous research has noted that DeepL Pro can translate a complete African anthology from English into French, leading to the belief that ChatGPT Pro would also possess the capability to translate an entire anthology. To identify and analyze the CSIs generated by the LLM, Aixelá's (1996) model was utilized. This model was previously applied in research (Opaluwah, 2024) to analyze CSIs in human translations (HT), General Machine Translations (GMT), and translations from a custom-built Microsoft Azure translation model.

## ChatGPT's Functionalities

This section mainly examines how ChatGPT's function has evolved and how it interprets and translates literary texts. Ogundare and Araya (2023), in the comparative Analysis of CHATGPT and the evolution of language models, utilized spontaneous quality (SQ) scores to compare the performance of ChatGPT on many NLP tasks such as machine translation, machine summarization, question answering, and language generation, compared with other mainstream algorithms. They propose that (LLMs) like ChatGPT have made many improvements in understanding and modelling language and bridging the gap between language models and knowledge models. With an increasing amount of available data for systematic training and the advancements in the computational theory of machine learning, it is expected that language models will continue to improve and push the boundaries of what is possible in natural language processing, especially applications in industrial automation.

Partha Pratim Ray (2023) comprehensively reviews ChatGPT's background, applications, key challenges, bias, ethics, limitations, and future scope in his work. Some of the issues he outlined that ChatGPT can work towards solving are maintaining context, handling ambiguity, personalization, common sense reasoning, emotional intelligence, ethical considerations, robustness and security, multi-modal interactions, out-of-distribution queries, scalability and efficiency. He concludes that over the years, LLMs have become larger and more powerful, with impressive natural language understanding and generation capabilities, and they have also been used in a wide range of natural language processing tasks, from language translation to question-answering systems. However, as these models become more complex, they also raise ethical and societal concerns, such as biases and limitations, which must be carefully addressed; their development requires substantial computational resources, including high-performance GPUs and extensive memory, and the training process often takes weeks or months.

In Servais Martial Akpaca's (2024) work on the Syntactic, Semantic, and Pragmatic Evaluation of the Translation of an Ethnographic Text by ChatGPT, the results of his study reveal that contrary to the current propaganda, ChatGPT primarily engages in literal translation. It does not engage in oblique translation. Indeed, errors and mistakes of syntactic, semantic, and pragmatic nature are abundant. Procedures such as transposition, modulation, foreignization, domestication, adaptation, transediting, etc., are almost unknown to it. According to him, ChatGPT is a tool containing many words and can effectively assist translators in their work. However, it is too early to envision a scenario where this technology would replace experienced translators. His evaluation involves identifying translation mistakes and subsequently revising the translation accordingly. After a questionnaire was submitted to ChatGPT, which it answered, he concluded that ChatGPT contains many words, but it is not the number of words a translator knows that makes them a good translator. This syntactic, semantic, and pragmatic evaluation of ChatGPT's translation has yielded disappointing results, leading to the conclusion that ChatGPT is merely a translation aid, even though it is more advanced than other tools such as Systran, Google Translate, Microsoft Translator, etc.

Shijing Si et al. (2023) explored ChatGPT's capabilities on ancient Chinese via two tasks: translating ancient Chinese to modern Chinese and recognizing ancient Chinese names. They evaluated the ability of ChatGPT on an ancient Chinese book, Shi Shuo Xin Yu, which is largely ignored by previous research. They also studied the performance of ChatGPT on ancient-to-modern translation by varying the input length for each query. Additionally, personal name recognition is rarely explored in ancient Chinese processing, given that ancient Chinese is challenging in natural language processing due to its complex grammatical structures, cultural nuances, and polysemy language. After designing proper prompt instructions for the translation task, it was discovered that the prompt rarely produced errors or superfluous information. Then, they extracted text content that contained only modern Chinese translations of ancient Chinese. They conclude that ChatGPT's capacity for ancient-to-modern Chinese translation and personal name recognition still leaves much to be desired. One possible explanation might be that ChatGPT is mainly pre-trained on English corpora with few Chinese corpora.

Yu Wei (2023), in his work on a comparative study between manual and ChatGPT Translations of Literary Texts, uses Kung I-chi as an Example in testing ChatGPT's literary translation ability at lexical, syntactic, and textual levels. While the research results showed that ChatGPT's literary translation has generally progressed, ChatGPT can hardly rival sophisticated human translators at the lexical level.

Faten Khoshafah (2023) also evaluated the accuracy of ChatGPT for Arabic-to-English translation. Her study focuses on different types of texts, including general texts from the media, literary, scientific, religious, and legal texts, to compare the accuracy of ChatGPT translation in these different contexts. From the results presented, it was discovered that the translations by ChatGPT and human translation carry the same meaning to some extent but with differences in style and word choice. Some texts translated by ChatGPT are grammatically correct but lack proper Arabic phrasing and word order. Idiomatic expressions are another challenge for ChatGPT. Arabic contains many colloquial terms that are not present in English. ChatGPT may also struggle with Arabic diacritical marks, which indicate the sounds of vowels and pronunciation. False cognates, words that appear or sound similar but have different meanings in both languages, can cause translation problems. Moreover, the rules, sentence structure, and cultural background differ between Arabic and English. She concludes that ChatGPT may not understand the cultural significance of some phrases or terms because it lacks the cultural background and context in which these expressions or terms were used.

Shanshan Wang et al. (2024) in Their study examine ChatGPT's capabilities in English-Chinese poetry translation tasks, utilizing targeted prompts and small sample scenarios to ascertain optimal performance. Despite promising outcomes, our analysis reveals persistent issues in the translations generated by ChatGPT that warrant attention. They used all 10 poems in the test set of the Explanation-Assisted Poetry Machine Translation (EAPMT*)* experiment to conduct data verification experiments on GPT-4 to verify the originality of ChatGPT's responses. They invited six judges to evaluate the experimental results of eight modern English poems translated under eight different prompts. These judges are native Chinese speakers and Chinese professional poets born between the 1960s and 1990s, who are members of the Chinese Writers Association who understand English, with some having engaged in poetry writing or English-Chinese poetry translation for many years. They asked the judges to vote on the output translations of GPT-3.5 and GPT-4 under three human prompts through a questionnaire survey to select the most appropriate prompt. At the end of their investigation, they observed that the decrease in the performance of the models GPT3.5 and GPT 4 was unexpected as more samples did not promote the models to learn the characteristics of poetry. The superior performance in the zero-shot scenario is likely due to the inherent nature of modern poetry, which is characterized by its freedom and lack of constraints, concluding that the optimum performance of GPT-3.5 on the poetry translation task occurs in the 0-shot scenario using the third prompt (H3). The best translation result is denoted as GPT3.5-Best. The optimum performance of GPT-4 occurs in the 0-shot scenario using the second prompt (H2), and the best translation result is recorded as GPT4-Best.

From the few works about ChatGPT's functions over time as LLM, it can be safely concluded that ChatGPT still has a lot to learn and is, therefore, still evolving in its understanding of more human languages and cultures.

## Methodology

This investigation aims to analyze the interpretation of CSIs by large language models (LLMs) using ChatGPT pro in the English-French translations of three African anthologies: *Early Poems, Mandela's Earth and Other Poems*, and *Idanre and Other Poems*. This analysis used structured prompts in tabular format, with a comparative assessment against human translations and a custom translation model. To achieve this, relevant sections from the anthologies, precisely two paragraphs above and below the CSIs, were copied and pasted into the chat following each prompt, and the outputs related to the CSIs were noted.

Four tables of comparisons were developed to facilitate this investigation. The initial table outlines the results of the CSIs generated after three distinct levels of prompts directed at ChatGPT Pro. The subsequent table categorizes the output of CSIs according to Aixelá's framework of "Proper nouns and Common expressions" (1996, p. 59), which will assist in evaluating the frequency of repetition within each category as produced by Human Translation (HT), custom-built Microsoft Azure translation engine (CTE), and LLM (ChatGPT Pro). The third table summarizes the outputs of CSIs from human translators, CTE, and LLM. Finally, a concluding table provides an overview of the strategies employed by ChatGPT, an LLM, in translating CSIs, particularly about Aixelá's guidelines, following the culture-oriented prompt.

Table 1
Overview of CSIs output by LLM

| N0. | Source word | General prompt: *Translate this poem into French* | Structure-oriented prompt: *Translate this poem to French, keeping the stanzas* | Culture-oriented prompt: *Translate this poem to French, paying attention to the cultural words* |
|---|---|---|---|---|
| 1. | My impi (Soyinka, 1976, p. 11) | Mon impi | Mon impi | Mon impi |
| 2. | Bean-cake (Soyinka, 1998, p. 133) x 2 | ruche de galettes de haricot<br><br>galette, marqués | ruche en galettes de haricots,<br><br>galettes gravés | ruche de gâteau de fève<br><br>le gâteau de fève |
| 3. | Kaffir (Soyinka, 1976, p. 20) | Kaffir | Kaffir | Kaffir |
| 4. | Sigidi (Soyinka, 1976, pp. 9, 11, 19, 22) | Sigidi | Sigidi | Sigidi |
| 5. | Bayete (Soyinka, 1976, pp. 10-14, 16) | Bayete | Bayete | Bayete |
| 6. | Ogun (Soyinka, 1976, pp. 9, 11, 19, 22)<br><br>(Soyinka, 1967, pp. 61, 63, 64, 65, 67-72, 74, 75, 80, 83, 85) | Ogun<br><br>Ogun | Ogun<br><br>Ogun | Ogun<br><br>Ogun |
| 7. | Abibimañ (Soyinka, 1976, pp. 5, 10, 22) | Abibimañ | Abibimañ | Abibimañ |
| 8. | Idanre (Soyinka, 1976, p. 10)<br><br>(Soyinka, 1967, pp. 68, 69, 72, 81, 82) | Idanre<br><br>Idanre | Idanre<br><br>Idanre | Idanre<br><br>Idanre |
| 9. | Esu (Soyinka, 1967, pp. 70 & 78) | Esu<br><br>*Esu* | Esu<br><br>*Esu* | Esu<br><br>*Esu* |
| | *Esu* (Soyinka, 1988, p. 48) | | | |
| 10. | Mfekane (Soyinka, 1976, p. 19 | Mfekane | Mfekane | Mfekane |
| 11. | Shaka (Soyinka, 1976, pp. 9, 10, 11-17) | Shaka | Shaka | Shaka |
| 12. | Mzilikazi (Soyinka, 1976, p. 12) | Mzilikazi | Mzilikazi | Mzilikazi |
| 13. | Soshangani (Soyinka, 1976, p. 12) | Soshangani | Soshangani | Soshangani |
| 14. | Lobengula (Soyinka, 1976, p. 12) | Lobengula | Lobengula | Lobengula |
| 15. | amaZulu (Soyinka, 1976, pp. 12, 14, 16, 17) | amaZulu | amaZulu | amaZulu |
| 16. | Sango (Soyinka, 1967, p. 61)<br><br>Sango (Soyinka, 1967, p. 70) | Sango<br><br>Sango | Sango<br><br>Sango | Sango<br><br>Sango |
| 17. | Atunda (Soyinka, 1967, pp. 81, 83) | Atunda | Atunda | Atunda |
| 18. | ogboni (Soyinka, 1967, p. 67) | ogboni | ogboni | ogboni |
| 19. | Ajantala (Soyinka, 1967, p. 67) | Ajantala | Ajantala | Ajantala |
| 20. | Orisa-nla (Soyinka, 1967, p. 70) | Orisa-nla | Orisa-nla | Orisa-nla |
| 21. | Orunmila (Soyinka, 1967, p. 70, 83) | Orunmila | Orunmila | Orunmila |
| 22. | Ifa (Soyinka, 1967, pp. 70, 83) | Ifa | Ifa | Ifa |
| 23. | Ire (Soyinka, 1967, p. 71) | Colère | Colère | Colère |
| 24. | Oya (Soyinka, 1967, p. 67) | Oya | Oya | Oya |
| 25. | Iron One (Soyinka, 1967, p. 61)<br><br>Iron One (Soyinka, 1967, pp. 68, 70, 74) | l'Homme de Fer<br><br>Le Fer, l'Homme de Fer, Ô Fer One | l'Homme de Fer<br><br>Le Fer, l'Homme de Fer, Ô Fer | le Fer<br><br>Le Fer, l'Iron One, Ô Fer, |
| | Iron One (Soyinka, 1967, p. 78) | l'Iron One | l'Acier | Fer One |
| 26 | Mickey Mouse (Soyinka, 1988, p. 41) | Mickey | Mickey Mouse | souris en peluche |
| 27 | Louiseville Lips (Soyinka, 1988, p. 47) | lèvres de Louisville | lèvres de Louisville | Louisville Lips |
| 28 | *Doppelgänger* (Soyinka, 1988, p. 10 | *Doppelgänger* | *Doppelgänger* | *Doppelgänger* |
| 29 | *kora* (Soyinka, 1988, p. 66) | *kora* | *kora* | *kora* |
| 30 | *gbegbe* (Soyinka, 1988, pp. 66 & 67) | *gbegbe* | *gbegbe* | *gbegbe* |
| 31 | *tete* (Soyinka, 1988, p. 67) | *tete* | *tete* | *tete* |
| 32 | *egungun* (Soyinka, 1988, p. 69) | *egungun* | *egungun* | *egungun* |
| 33 | *Akaraba* (Soyinka, 1988, p. 50) | *Akaraba* | *Akaraba* | *Akaraba* |
| 34 | *jigida* (Soyinka, 1988, p. 45) | *jigida* | *jigida* | *jigida* |
| 35 | *kakaki* (Soyinka, 1988, p. 66) | *kakaki* | *kakaki* | *kakaki* |
| 36 | *mbira* (Soyinka, 1988, p. 48) | *mbira* | *mbira* | *mbira* |
| 37 | *Ile-Ife* (Soyinka, 1988, p. 61) | *Ile-Ife* | *Ile-Ife* | *Ile-Ife* |
| 38 | Mandel (Soyinka, 1988, p. 6) | Mandel | Mandel | Mandel |
| 39 | Mendel (Soyinka, 1988, p. 6) | Mendel | Mendel | Mendel |
| 40 | Mengel (Soyinka, 1988, p. 6) | Mengel | Mengel | Mengel |
| 41 | Mengele (Soyinka, 1988, p. 6) | Mengele | Mengele | Mengele |
| 42 | Broederland (Soyinka, 1988, p. 8) | Broederland | Broederland | Broederland |
| 43 | Mandgela (Soyinka, 1988, p. 8) | Mandgela | Mandgela | Mandgela |
| 44 | Mendgela (Soyinka, 1988, p. 9) | Mendgela | Mendgela | Mendgela |
| 45 | Biko (Soyinka, 1988, p. 8) | *Biko* | *Biko* | *Biko* |
| 46. | Scottsborough Boys (Soyinka, 1988, p. 40) | Scottsborough Boys | Scottsborough Boys | Scottsboro Boys |
| 47. | Recolored Brutus. (Soyinka, 1988, p. 9) | re-colored Brutus, | re-colored Brutus, | re coloré Brutus |
| 48. | Mister Boots, Knucles and Bones (Soyinka, 1988, p. 9) | Monsieur Bottes, Articulations et Os | Monsieur Bottes, Coudes et Os | Monsieur Boots, Knuckles et Bones. |
| 49. | Salem Seers (Soyinka, 1988. P. 13) | Voyants de Salem | Voyants de Salem | Voyants de Salem |
| 50. | Ceasar (Soyinka, 1988, p. 19 & 31) | César | César | César |
| 51. | Swinging Bokassa (Soyinka, 1988, p. 33 | Balançant Bokassa | Swinguant Bokassa | Balançant Bokassa |
| 52. | Master Sergent Doe (Soyinka, 1988, p. 33) | Sergent Major Doe | Sergent Maître Doe | Sergent Maître Doe |
| 53. | *Asantehene* (Soyinka, 1988. P. 17) | *Asantehene* | *Asantehene* | *Asantehene* |
| 54. | *sjambok* (Soyinka, 1988 , p. 3) | *sjambok* | *sjambok* | *sjambok* |

**Unchanged (Repeated) CSIs**

With all three types of prompts, 41 CSIs out of 54 are unchanged: *My impi*, *Kaffir*, *Sigidi*, *Bayete*, *Ogun*, *Abibimañ*, *Idanre*, *Esu*, *Mfekane*, *Shaka*, *Mzilikazi*, *Soshangani*, *Lobengula*, *amaZulu*, *Sango*, *Atunda*, *ogboni*, *Ajantala*, *Orisa-nla*, *Orunmila*, *Ifa*, *Oya*, *Doppelgänger, kora*, *gbegbe*, *tete*, *egungun*, *Akaraba*, *jigida*, *kakak*i, *mbira*, *Ile-Ife*, *Mandel*, Mendel, *Mengel*, *Mengele*, *Broederland*, *Mandgela*, *Mendgela*, *Biko*, *Asantehene* and *sjambok* were rendered by ChatGPT, unchanged. In other words, the above CSIs were rendered through the strategy of "Repetition" (Aixelá, 1996, p. 61).

To categorize the above CSIs, Aixelá's model distinguishes

> "Two basic categories from the translator's point of view: proper nouns and common expressions (for want of a better term, the world of objects, institutions, habits, and opinions restricted to each culture and that cannot be included in the field of proper names)" (1996, p. 59)

Thus, the 41 CSIs that have retained their forms as in the source texts are grouped below, giving 27 proper nouns and 14 common expressions.

Table 2
Overview of Unchanged CSIs after the three levels of prompts

| Proper Nouns | Common Expressions |
|---|---|
| Kaffir | My impi |
| Ogun | Bayete |
| Abibimañ | Mfekane |
| Idanre | Sigidi |
| Esu | ogboni |
| Shaka | kora |
| Mzilikazi | *Broederland* |
| Soshangani | jigida |
| Lobengula | mbira |
| Atunda | gbegbe |
| Ajantala | kakaki |
| Orisa-nla | egungun |
| Orunmila | tete |
| Ifa | sjambok |
| Oya | |
| Ile-Ife | |
| Mandel | |
| Mendel | |
| Mengel | |
| Mengele | |
| Broederland | |
| Doppelgänger | |
| Akaraba | |
| Mandgela | |
| Mendgela | |
| Biko | |
| Asantehene | |

Table 3
Overview of CSIs output by ChatGPT, HT and CTE

| | Source word | Human translation | ChatGPT translation after the culture-oriented prompt. | CTE output of CSIs. |
|---|---|---|---|---|
| 1. | My impi (Soyinka, 1976, p. 11) | Mes impi (Soyinka 2019:160) | Mon impi | Mes *impi* |
| 2. | Bean-cake (Soyinka, 1998, p. 133) x 2 | Gâteau de fèves (Soyinka 2019, p. 77)<br><br>Gâteau de fèves (Soyinka 2019, p. 77) | ruche de gâteau de fève<br><br>le gâteau de fève | gâteau de fèves<br><br>gâteau de fèves |
| ***3.*** | Kaffir (Soyinka, 1976, p. 20) | ------ | Kaffir | Kaffir |
| ***4.*** | Sigidi (Soyinka, 1976, pp. 9, 11, 19, 22) | Sigidi (Soyinka 2019, pp. 157, 158, 160, 169, 173, 175, 177) | Sigidi | Sigidi |
| ***5.*** | Bayete (Soyinka, 1976, pp. 10-14, 16) | Bayete (Soyinka 2019, pp. 158, 160-163, 165, 167, 168, 176, 177) | Bayete | Bayete |
| ***6.*** | Ogun (Soyinka, 1976, pp. 9, 11, 19, 22)<br><br>(Soyinka, 1967, pp. 61, 63, 64, 65, 67-72, 74, 75, 80, 83, 85) | Ogun (Soyinka 2019, pp. 151, 154, 156, 158, 160, 170, 173-178)<br><br>Ogoun Soyinka, 1982, p. 7, 8, 10, 12, 15, 17, 19-21, 23, 26, 27, 28, 32 | Ogun<br><br>Ogun | Ogun<br><br>Ogoun |
| ***7.*** | Abibimañ (Soyinka, 1976, pp. 5, 10, 22) | Abibimañ (Soyinka 2019, pp. 149, 154, 159, 175, 177) | Abibimañ | Abibimañ |
| ***8.*** | Idanre (Soyinka, 1976, p. 10)<br><br>(Soyinka, 1967, pp. 68, 69, 72, 81, 82) | Idanré (Soyinka 2019, p. 158)<br><br>Idanré (Soyinka, 1989, p. 91) | Idanre<br><br>Idanre | Idanre,<br><br>Idanré |
| ***9.*** | Esu (Soyinka, 1967, pp. 70 & 78)<br><br>*Esu* (Soyinka, 1988, p. 48) | ………..<br><br>Esu (Soyinka, 1982, pp. 27 & 34 | Esu<br><br>*Esu* | Eshou<br><br>Esu |
| ***10.*** | Mfekane (Soyinka, 1976, p. 19 | Mfekane (Soyinka 2019, pp. 157, 169 | Mfekane | Mfekane |
| ***11.*** | Shaka (Soyinka, 1976, pp. 9, 10, 11-17) | ………. | Shaka | Chaka |
| ***12.*** | Mzilikazi (Soyinka, 1976, p. 12) | Mzilikazi (Soyinka 2019, p. 162) | Mzilikazi | Mzilikazi |
| ***13.*** | Soshangani (Soyinka, 1976, p. 12) | Soshangani (Soyinka 2019, p. 162) | Soshangani | Soshangani |
| ***14.*** | Lobengula (Soyinka, 1976, p. 12) | Lobengula (Soyinka 2019, p. 162) | Lobengula | Lobengula |
| ***15.*** | amaZulu (Soyinka, 1976, pp. 12, 14, 16, 17) | ------------- | amaZulu | amaZulu |
| ***16.*** | Sango (Soyinka, 1967, p. 61)<br><br>Sango (Soyinka, 1967, p. 70) | Chango (Soyinka, 1982, p. 17)<br><br>Shango (Soyinka, 1982, p. 27) | Sango<br><br>Sango | Shango,<br><br>Sango,<br><br>Chango |
| ***17.*** | Atunda (Soyinka, 1967, pp. 81, 83) | Atunda (Soyinka, 1982, p. 37 & 38) | Atunda | Atunda |
| ***18.*** | ogboni (Soyinka, 1967, p. 67) | Ogboni (Soyinka, 1982, p. 22) | ogboni | ogboni |
| ***19.*** | Ajantala (Soyinka, 1967, p. 67) | Ajantala (Soyinka, 1982, p. 22) | Ajantala | Ajantala |
| ***20.*** | Orisa-nla (Soyinka, 1967, p. 70) | Orisanla (Soyinka, 1982, p. 27) | Orisa-nla | Orisa-nla |
| ***21.*** | Orunmila (Soyinka, 1967, p. 70, 83) | Orunmila (Soyinka, 1982, p. 27 & 38) | Orunmila | Orunmila |
| ***22.*** | Ifa (Soyinka, 1967, pp. 70, 83) | Ifa (Soyinka, 1982, p. 27 & 38) | Ifa | Ifa |
| ***23.*** | Ire (Soyinka, 1967, p. 71) | Ire (Soyinka, 1982, p. 27) | Colère | Ire |
| ***24.*** | Oya (Soyinka, 1967, p. 67) | Oya (Soyinka, 1982, p. 22) | Oya | Oya |
| ***25.*** | Iron One (Soyinka, 1967, p. 61)<br><br>Iron One (Soyinka, 1967, pp. 68, 70, 74)<br><br>Iron One (Soyinka, 1967, p. 78) | Fer en personne (Soyinka, 1982, p. 17)<br><br>dieu du fer (Soyinka, 1982, pp. 25, 27, 32)<br><br>Ogoun (Soyinka, 1982, p. 34) | le Fer<br><br>Le Fer, l'Iron One, Ô Fer,<br><br>Fer One | Fer en personne<br><br>Dieu du fer, Ogoun<br><br>Dieu de fer |
| ***26.*** | Mickey Mouse (Soyinka, 1988, p. 41) | Mickey Mouse (Soyinka, 1989, p. 63) | souris en peluche | Mickey Mouse |
| ***27.*** | Louiseville Lips (Soyinka, 1988, p. 47) | à Louisville (Soyinka, 1989, p. 71) | Louisville Lips | à Louisville |
| ***28.*** | *Doppelgänger* (Soyinka, 1988, p. 10 | *Doppelgänger* (Soyinka, 1989, p. 22) | *Doppelgänger* | Doppelgänger |
| ***29.*** | *kora* (Soyinka, 1988, p. 66) | *Kora* (Soyinka, 1989, p. 99) | *kora* | kora |
| ***30.*** | *gbegbe* (Soyinka, 1988, pp. 66 & 67) | *Gbegbe* (Soyinka, 1989, pp. 99 & 100) | *gbegbe* | gbegbe |
| ***31.*** | *tete* (Soyinka, 1988, p. 67) | *Tete* (Soyinka, 1989, p. 100) | *tete* | tete |
| ***32.*** | *egungun* (Soyinka, 1988, p. 69) | *egungun* (Soyinka, 1989, p. 103) | *egungun* | egungun |
| ***33.*** | *Akaraba* (Soyinka, 1988, p. 50) | *Akaraba* (Soyinka, 1989, p. 75) | *Akaraba* | Akaraba |
| ***34.*** | *jigida* (Soyinka, 1988, p. 45) | *jiguida* (Soyinka, 1989, p. 68) | *jigida* | jigida |
| ***35.*** | *kakaki* (Soyinka, 1988, p. 66) | *kakaki* (Soyinka, 1989, p. 99) | *kakaki* | kakaki |
| ***36.*** | *mbira* (Soyinka, 1988, p. 48) | *mbira* (Soyinka, 1989, p.72) | *mbira* | mbira |
| ***37.*** | *Ile-Ife* (Soyinka, 1988, p. 61) | *Ilé-Ifé* (Soyinka, 1989, p. 91) | *Ile-Ife* | Ile-Ife |
| ***38.*** | Mandel (Soyinka, 1988, p. 6) | Mandel (Soyinka, 1989, p.17) | Mandel | Mandel |
| ***39.*** | Mendel (Soyinka, 1988, p. 6) | Mendel (Soyinka, 1989, p.17) | Mendel | Mendel |
| ***40.*** | Mengel (Soyinka, 1988, p. 6) | Mengel (Soyinka, 1989, p.17) | Mengel | Mengel |
| ***41.*** | Mengele (Soyinka, 1988, p. 6) | Mengelle (Soyinka, 1989, p.17) | Mengele | Mengele |
| ***42.*** | Broederland (Soyinka, 1988, p. 8) | Broederland (Soyinka, 1989, p. 20) | Broederland | Broederland |
| ***43.*** | Mandgela (Soyinka, 1988, p. 8) | Mandgela (Soyinka, 1989, p. 20) | Mandgela | Mandgela |
| ***44.*** | Mendgela (Soyinka, 1988, p. 9) | Mandgela (Soyinka, 1989, p. 21) | Mendgela | Mendgela |
| ***45.*** | Biko (Soyinka, 1988, p. 8) | Biko (Soyinka, 1989, p. 20) | *Biko* | Biko |
| ***46.*** | Scottsborough Boys (Soyinka, 1988, p. 40) | Enfants de Scottsborough (Soyinka, 1989, p. 62) | Scottsboro Boys | les enfants de Scottsborough |
| ***47.*** | Recolored Brutus. (Soyinka, 1988, p. 9) | ----------------- | re coloré Brutus, | Brutus recoloré |
| ***48.*** | Mister Boots, Knucles and Bones (Soyinka, 1988, p. 9) | Monsieur Bottes, Jointures et Os (Soyinka, 1989, p. 22) | Monsieur Boots, Knuckles et Bones. | Monsieur Bottes, Jointures et Os |
| ***49.*** | Salem Seers (Soyinka, 1988. P. 13) | Voyantes de Salem (Soyinka, 1989, p. 26) | Voyants de Salem | Voyantes de Salem |
| ***50.*** | Ceasar (Soyinka, 1988, p. 19 & 31) | --------------- | César | César |
| ***51.*** | Swinging Bokassa (Soyinka, 1988, p. 33 | --------------- | Balançant Bokassa | fringants Bokassa |
| ***52.*** | Master Sergent Doe (Soyinka, 1988, p. 33) | --------------- | Sergent Maître Doe | sergent-chef Doe |
| ***53.*** | *Asantehene* (Soyinka, 1988. P. 17) | --------------- | *Asantehene* | Ashantihini |
| ***54.*** | *sjambok* (Soyinka, 1988 , p. 3) | *sjambok* (Soyinka, 1989, p. 12) | *sjambok* | Sjambok |

CSIs translation strategies by LLM, following Aixelá's (1996) model

Table 4
CSIs Translation Strategies by ChatGPT Pro for changed CSIs

| Source word | CSIs | And | Corresponding | Strategies | After each | prompt |
|---|---|---|---|---|---|---|
| | General Prompt | Strategy | Structure-oriented prompt | Strategy | Culture-oriented prompt | Strategy |
| Bean-cake | ruche de galettes de haricot<br><br>galette, marqués | Linguistic (non-cultural) translation<br><br>Linguistic (non-cultural) translation | ruche en galettes de haricots,<br><br>galettes gravés | Linguistic (non-cultural) translation<br><br>Linguistic (non-cultural) translation | ruche de gâteau de fève<br><br>le gâteau de fève | Linguistic (non-cultural) translation<br><br>Linguistic (non-cultural) translation |
| Ire | Colère | Autonomous creation | Colère | Autonomous creation | Colère | Autonomous creation |
| Iron One | l'Homme de Fer | Linguistic (non- | l'Homme de Fer | Linguistic (non- | le Fer | Absolute Universalization |
| | Le Fer,<br><br>l'Homme de Fer,<br><br>Ô Fer One<br><br>l'Iron One | cultural) translation<br><br>absolute Universalization,<br><br>Linguistic (non-cultural) translation,<br><br>limited Universalization<br><br>Repetition | Le Fer,<br><br>l'Homme de Fer,<br><br>Ô Fer<br><br>l'Acier | cultural) translation<br><br>Absolute Universalization,<br><br>Linguistic (non-cultural) translation,<br><br>Absolute Universalization<br><br>Absolute Universalization | Le Fer,<br><br>l'Iron One,<br><br>Ô Fer,<br><br>Fer One | Absolute Universalization,<br><br>Repetition,<br><br>Absolute Universalization<br><br>Linguistic (non-cultural) translation |
| Mickey Mouse | ------------- | ------------- | Mickey Mouse | Repetition | Souris en peluche | Linguistic (non-cultural) translation |
| Louisville Lips | lèvres de Louisville | Linguistic (non-cultural) translation | lèvres de Louisville | Linguistic (non-cultural) translation | Louisville Lips | Repetition |
| Scottsborough Boys | Scottsborough Boys | Repetition | Scottsborough Boys | Repetition | Scottsboro Boys | Limited Universalization |
| Recolored Brutus | Re-colored Brutus | Repetition | Re-colored Brutus | Repetition | re coloré Brutus | Linguistic (non-cultural) translation |
| Mister Boots, Knuckles and Bones | Monsieur Bottes, Articulations et Os | Linguistic (non-cultural) translation | Monsieur Bottes, Coudes et Os | Linguistic (non-cultural) translation | Mister Boots, Knuckles et Bones | Linguistic (non-cultural) translation |
| Salem Seers | Voyants de Salem | Linguistic (non-cultural) translation | Voyants de Salem | Linguistic (non-cultural) translation | Voyants de Salem | Linguistic (non-cultural) translation |
| Swinging Bokassa | Balançant Bokassa | Linguistic (non-cultural) translation | Swiguant Bokassa | Linguistic (non-cultural) translation | Balançant Bokassa | Linguistic (non-cultural) translation |
| Master Sergent Doe | Sergent Major Doe | Linguistic (non-cultural) translation | Sergent Maître Doe | Linguistic (non-cultural) translation | Sergent Maître Doe | Linguistic (non-cultural) translation |

## Analysis

Tables 1 and 2 indicate that the LLM produced 41 CSIs in repetition, translating them precisely as they appear in the source text. Among these, 27 are classified as Proper Nouns, while 14 are identified as Common Expressions. In a prior study, CTE repeated 38 CSIs in their original form from the source anthology, comprising 25 Proper Nouns and 13 Common Expressions. In contrast, the reference (human) translation preserved 33 CSIs unchanged, with 21 categorized as Proper Nouns and 12 as Common Expressions.

An examination of how the language model translated certain CSIs that were neither repeated nor left untranslated reveals inconsistencies in its approach to translating African poetry from English to French following three distinct levels of prompts. For instance, after the initial general prompt, ChatGPT translated "Bean-cake" as "ruche de galettes de haricot" and "galette" in the subsequent translation. Following the structure-oriented prompt, the same term was rendered as "ruche en galettes de haricots" and "galettes gravés" in the second instance. The first notable difference in the output after the second prompt pertains to using prepositions (en, de). In contrast, the second difference lies in selecting adjectives (marqués, gravés). After the culture-oriented prompt, ChatGPT provided "ruche de gâteau de fève" in the first instance and "le gâteau de fève" in the second.

The term "Ire" was translated as "Colère" following three prompts. Ire refers to "a town in Ekiti" (Scheub, 2000, Kindle Location 3238), yet ChatGPT has transformed this place noun into an abstract noun. In the case of "Iron One," five different translations was produced including "L'homme de Fer," "Le Fer," "L'homme de Fer," "Ô Fer," and "L'Iron One" after the initial prompt. The second prompt yielded "L'homme de Fer," "Le Fer," "L'homme de Fer," "Ô Fer," and "l'Acier." Following the culture-specific prompt, the translations for "Iron One" were "Le Fer," "Le Fer," "l'Iron One," "O Fer," and "Fer One." The most frequently occurring translations are "L'homme de Fer" and "Le Fer." Notably, even after the culture-oriented prompt, ChatGPT's translation of "Iron One" does not align closely with the reference human translations such as "Fer en personne" (Soyinka, 1982, p. 17), "dieu du fer" (Soyinka, 1982, pp. 25, 27, 32), and "Ogoun" (Soyinka, 1982, p. 34). Two other cases to note are LLM's translation of "Mickey Mouse" as "souris en peluche" and "Scottsborough Boy"s as "Scottsboro Boys" (without the **ugh**) after the culture-oriented prompt.

As illustrated in Table 4, ChatGPT's five strategies for the translation of cultural-specific items (CSIs), as per Aixela's (1996) framework, encompass Repetition, Linguistic (non) cultural translation, Autonomous creation, Absolute Universalization, and Limited Universalization. In a previous study, the strategies associated with CTE included Repetition, Linguistic (non) cultural translation, Absolute Universalization, Limited Universalization, and Synonymy (Opaluwah, 2024). In contrast, Human translators adopted only four strategies: Repetition, Linguistic (non) cultural translation, Limited Universalization, and Synonymy.

## Conclusion

The analysis demonstrates that culture-oriented prompts applied to ChatGPT pro, an LLM, did not result in any meaningful differences in the output of culture-specific items in the translation of African poetry from English to French. Specifically, in translating "Iron One," the model suggested "L'homme de Fer" as a translation option after utilizing general, structural, and culture-oriented prompts. Still, the culture-oriented prompt did not significantly improve the translation quality. Moreover, the translation of "Scottsborough Boys" yielded a less acceptable result, being rendered as "Scottsboro Boys" without the necessary "ugh" in the word. In the case of "Mickey Mouse," the culture-oriented prompt led to the translation "souris en peluche".

While LLMs have shown remarkable potential, they also present challenges such as inherent biases, ethical concerns, and interpretability issues due to their size and complexity. These factors can limit their application in some areas, highlighting the need for responsible development and deployment. Despite these concerns, LLMs continue to evolve, becoming more powerful and capable, with an expanding role in enhancing human-computer interaction. Their development trajectory has seen the creation of increasingly sophisticated models, enabling advanced NLP tasks like language translation and question-answering systems. However, ethical and societal concerns regarding their use require careful consideration to ensure fairness and mitigate biases. As research progresses, LLMs are expected to become even more integral to various industries and aspects of daily life (Ray, 2024).

A comparison of the CSI output from a custom-trained engine for translating African poetry from English to French reveals that CTE rendered superior CSIs compared to LLMs, even when culture-oriented prompts are employed for LLM. This conclusion is consistent with the findings of Kuzman et al. (2019), which argue that models customized for specific authors (the CTE was customized to Wole Soyinka's poetry) generally outperform those developed using a more extensive and varied literary corpus. Additional investigation is required to understand how LLMs interpret prompts across various literary domains like drama, social media, etc, in different language pairs from diverse regions.

## References

Aixelá, J. F. (1996). Culture-Specific Items in Translation. In R. Alvarez and M. Vidal (Eds.), *Translation, Power, Subversion* (52-78). Multilingual Matters.

Akpaca, S. M. (2024). A Syntactic, Semantic, and Pragmatic Evaluation of the Translation of an Ethnographic Text by ChatGPT. *London Journal of Research in Humanities & Social Science, 24*(12). Retrieved from https://journalspress.com/LJRHSS_Volume24/A-Syntactic-Semantic-and-Pragmatic-Evaluation-of-the-Translation-of-an-Ethnographic-Text-by-ChatGPT.pdf

Khoshafah, F. (2023). ChatGPT for Arabic-English Translation: Evaluating the Accuracy. *Research square*. Retrieved from https://assets-eu.researchsquare.com/files/rs-2814154/v1_covered.pdf

Kuzman, T. et al. (2019). *Neural Machine Translation of Literary Texts from English to Slovene.* [Conference Presentation]. Dublin Machine Translation Summit Aug 2019. W19-73.pdf (aclweb.org)

Liebman, S. (2018). *How to Say Mickey Mouse in 27 Different Languages.* Disney Diary. Retrieved from https://disneydiary.com/2018/09/how-to-say-mickey-mouse-in-27-different-languages/

Ogundare, O. & Araya, G. Q. (2023). Comparative Analysis of CHATGPT and the Evolution of Language Models. *Arxiv*. Retrieved from https://arxiv.org/pdf/2304.02468

Opaluwah, A. (2025). A Case for Machine in The Translation of Culture-Specific Items. In Sara García Fernández, Fátima Gómez-Cáneba, Beatriz Guerrero García, Francesca Placidi, Margarita Savchenkova, Susana Schoer-Granado (Eds.), *Traducción y sostenibilidad cultural II: retos y nuevos escenarios* (pp. 659- 671).

Ediciones Universidad Salamanca. https://doi.org/10.14201/0AQ0373659671

Ray, P.P. (2023). ChatGPT: A comprehensive review of background, applications, key challenges, bias, ethics, limitations and future scope. *ScienceDirect, 3,* 121-154. Retrieved from https://www.sciencedirect.com/science/article/pii/S266734522300024X

Scheub, H. (2000). *A Dictionary of African Mythology: The Mythmaker as Storyteller*. New York, USA: Oxford University Press. Kindle Edition.

Si, S. et al. (2023). Evaluating the Capability of ChatGPT on Ancient Chinese. *Arxiv*. Retrieved from https://arxiv.org/html/2312.15304v1

Soyinka, W. (1967). *Idanre and Other Poems*. New York, USA: Hill and Wang.

Soyinka, W. (1976). *Ogun Abibimañ*. London, United Kingdom: Rex Collings.

Soyinka, W. (1982). *Idanre: poème*. (A. Bordeaux, Trans.). Dakar : Nouvelles éditions Africaines. (Original work published 1967)

Soyinka, W. (2019). *Cycles sombres*. (E. Galle, Trans.). Coédition NENA/Panafrika/Silex/Nouvelles du Sud. Kindle Edition.

Soyinka, W. (1998). *Early Poems*. New York: Oxford University Press.

Wang, S. et al (2024). What is the Best way for ChatGPT to Translate Poetry? *Arxiv*. Retrieved from https://arxiv.org/abs/2406.03450

Wei, Y. (2023). A Comparative Study Between Manual and ChatGPT Translations on Literary Texts Taking Kung I-chi as an Example. *Asia-Pacific Journal of Humanities and Social*

*Sciences, 3*(4), *57-66*. Retrieved from http://www.appublishing.com.au/public/upload/file/20231219/1702971274160762.pdf